\documentclass[conference]{IEEEtran}
\IEEEoverridecommandlockouts

\usepackage{cite}
\usepackage{graphicx}
\usepackage{booktabs}
\usepackage{amsmath,amssymb,amsfonts}
\usepackage{url}
\usepackage{xcolor}
\usepackage{tikz}
\usetikzlibrary{arrows.meta}
\usetikzlibrary{positioning}
\usepackage{algorithm}
\usepackage{algorithmic}

\usepackage[caption=false,font=footnotesize]{subfig}

\begin{document}

\title{Improving LLM performance through black-box online tuning:
a case for adding system specs to Factsheets for Trusted AI
}

\author{\IEEEauthorblockN{1\textsuperscript{st} Yonas Atinafu}
\IEEEauthorblockA{\textit{David R. Cheriton School of Computer Science} \\
\textit{University of Waterloo}\\
Waterloo, Canada \\
yatinafu@uwaterloo.ca}
\and
\IEEEauthorblockN{2\textsuperscript{nd} Henry Lin}
\IEEEauthorblockA{\textit{David R. Cheriton School of Computer Science} \\
\textit{University of Waterloo}\\
Waterloo, Canada \\
h293lin@uwaterloo.ca}
\and
\IEEEauthorblockN{3\textsuperscript{rd} Robin Cohen}
\IEEEauthorblockA{\textit{David R. Cheriton School of Computer Science} \\
\textit{University of Waterloo}\\
Waterloo, Canada \\
rcohen@uwaterloo.ca}
}

\maketitle
\thispagestyle{empty}

\begin{abstract}
In this paper, we present a novel approach for a black-box online controller that uses only end-to-end measurements over short segments (no internal instrumentation) and hill-climbing to maximize goodput (the throughput of requests that satisfy the service-level objective). We provide empirical evidence to suggest that the design is well-founded. With this specific advance in LLM serving on hand as a clear example, we move on to discuss the importance of integrating system performance and sustainability metrics into Factsheets for those adopting AI systems.
\end{abstract}



\section{Introduction}
\label{sec:introduction}
In this paper we introduce a challenge to LLM throughput and user-facing performance in deployment: interactive services are dominated by tail latency~\cite{dean2013tail}, and practical operating points are governed by a small set of serving-stack settings (client concurrency, batching limits, and speculative-decoding parameters). Because these settings interact in workload- and hardware-dependent ways, default configurations can either underutilize expensive GPUs or push the system past a queueing challenge causing p99 to spike and a minority of users to experience extreme delays. We then present a novel approach for addressing this problem: SLO-Tuner\footnote{Code and artifacts:\url{https://github.com/Seoulsim2/SLO-Tuner}.}, a black-box online controller that uses only end-to-end measurements over short segments (no internal instrumentation)and a hill-climb aware of the p99 value (indicating when one percent of requests for service may fail under load) to maximize goodput(the throughput of requests that satisfy the service-level objective). Here the SLO (service-level objective) is an explicit tail-latency target, e.g., p99 $\le$ 1.2~s. We provide empirical evidence to suggest that the design is well-founded: on TinyLlama served with vLLM, SLO-Tuner improves a default-like configuration from $\approx$1.36\,s to $\approx$0.70\,s p99 while increasing goodput from $\approx$8 to $\approx$15 requests/s under a 1.2\,s SLO. To support efficient exploration and stress-testing, we also introduce a lightweight discrete-event simulator that captures dominant queueing/batching dynamics and reproduces the same qualitative trends as the live system.
With this insight into concerns with LLM performance, we discuss the importance of Responsible AI and then argue that it is both necessary and feasible for user factsheets to integrate additional performance metrics (including tail-latency SLO compliance) in order to support Trusted AI. We argue that deployment realities shape reliability, fairness at the tail and adoption in practice, also advocating for sustainability metrics.

\noindent\textbf{Contributions.}
\textbf{[1] SLO-first objective:} We formulate online tuning for LLM serving as maximizing \emph{goodput} (SLO-satisfying throughput) under an explicit p99 SLO, rather than optimizing average throughput/latency.
\textbf{[2] Speculation as runtime control:} We treat speculative decoding \emph{parameters} as tunable runtime settings whose SLO-safe values are workload-dependent and can be counterproductive for p99.
\textbf{[3] Portable logical knobs:} We introduce a small set of operator-facing \emph{logical knobs} (queueing pressure, batch formation, speculation aggressiveness) plus a thin adapter that maps them to stack-specific flags, enabling black-box deployment.
\textbf{[4] Simulator--live alignment:} We provide a discrete-event simulator that captures the dominant dynamics and aligns qualitatively with live vLLM behavior, supporting simulator-guided search and stress-testing.

\section{Tuning LLM servers towards improved throughput and fairness} \label{sec:tuning}
We study the practical problem of configuring an LLM \emph{serving stack}: a system that receives user prompts and returns generated text under real-time constraints. In deployment, operators face a familiar tension: pushing for higher GPU utilization (via more parallel requests and larger batches) can sharply increase queueing delays for some users. Because users experience the \emph{worst} delays disproportionately, we focus on \emph{tail latency} (p99) and treat meeting a p99 service-level objective (SLO) as a first-class constraint while maximizing throughput. 

\noindent\textbf{Background:}
\textbf{[1] SLOs and goodput.} Interactive LLM services are judged by \emph{tail latency}, so we target a p99 service-level objective (SLO), e.g., p99 $\le$ 1.2~s (99\% of requests complete within 1.2 seconds)~\cite{googleSRE_slos}. A p99 SLO is a fairness constraint: it prevents a minority of users from suffering extreme delays while averages look fine. Because raw throughput can look high even when many requests violate the SLO, we optimize \emph{goodput}: the rate of requests that \emph{meet} the SLO (violations contribute zero goodput).
\textbf{[2] Server knobs and the ``knee.''} Goodput and p99 are shaped by a small set of server knobs: (i) client concurrency (requests in flight), (ii) batching (sequences grouped for GPU execution, e.g., \texttt{max\_num\_seqs}), and (iii) speculative decoding parameters. Increasing concurrency or batch size can raise GPU utilization but also increases queueing, often producing a sharp ``knee'' where p99 spikes. Speculative decoding~\cite{leviathan2023speculative} uses a smaller draft model to propose tokens that a larger target model verifies; it can reduce average latency, but rejected drafts add verification work and variance that can harm p99. Current stacks expose speculative settings as runtime flags but do not tune them automatically for a target p99 SLO~\cite{vllm_docs_spec_decode}.

\subsection{Related work}
\textbf{Black-box tuning under tail constraints.} Prior systems treat configuration as black-box optimization (e.g., OtterTune~\cite{chen2019ottertune}, CherryPick~\cite{alipourfard2017cherrypick}). Our setting differs in what must be safe and what matters: LLM serving is dominated by tail latency~\cite{dean2013tail}, so we optimize \emph{goodput} under an explicit p99 SLO using only end-to-end measurements (no instrumentation), and do so \emph{online} near the SLO boundary. 
\textbf{Inference-engine tuning for SLOs:} SCOOT~\cite{cheng2025scoot} optimizes LLM-serving SLOs by Bayesian optimization over inference-engine parameters under stress testing, learning hidden feasibility constraints with a random forest. Unlike SCOOT (which excludes speculative decoding), we treat speculative decoding as a first-class runtime knob and favor a deterministic neighbor hill-climb plus a co-designed discrete-event simulator for safe pre-deployment exploration and trend validation.
\textbf{QoS vs.\ intra-replica control:} QoS-aware systems such as Quasar~\cite{delimitrou2014quasar} and Sinan~\cite{gan2021sinan} primarily act through cluster scheduling/placement. We instead target \emph{intra-replica} knobs (concurrency, batching, and speculative decoding), which can violate p99 even with good cluster-level decisions. Our contribution is to expose a portable, operator-facing formulation of this local control problem (goodput under p99 constraint) and a black-box controller that can be deployed alongside existing cluster schedulers and autoscalers rather than replacing them. \textbf{Speculation as a runtime knob:} Speculative decoding improves average decoding speed~\cite{leviathan2023speculative} and has many algorithmic variants (e.g., Medusa/Hydra~\cite{cai2024medusa,ankner2024hydra}); we treat speculation as a \emph{runtime control surface} whose best setting is workload- and SLO-dependent (as exposed in serving stacks such as vLLM~\cite{vllm_docs_spec_decode}).  \textbf{Simulator-guided, measurement-correct tuning:} We use a lightweight discrete-event simulator for cheap stress-testing and trend guidance, while final decisions come from online measurements on the target hardware. This division of labor---simulator for guidance, online measurements for correctness---supports practical deployment while preserving portability across serving stacks and accelerators.

\subsection{SLO-Tuner system design}
\begin{figure}[t]
  \centering
  \begin{tikzpicture}[
      >=Latex,
      font=\scriptsize,
      node distance=8mm and 8mm,
      box/.style={
        draw,
        rounded corners=1pt,
        fill=gray!12,
        align=center,
        text width=0.25\columnwidth,
        inner sep=2pt
      },
      arr/.style={-Latex, thick}
    ]
    \node[box] (gen) {Request generator\\(prompts, output cap)};
    \node[box, right=of gen] (serv) {Simulator / vLLM\\FCFS batching\\speculative decode};
    \node[box, right=of serv] (met) {Metrics\\p50/p95/p99, goodput};
    \node[box, below=of serv] (tun) {SLO-Tuner\\hill-climb controller\\knobs: concurrency,\\batch size, spec width};

    \draw[arr] (gen) -- (serv);
    \draw[arr] (serv) -- (met);
    \draw[arr] (met.south) |- (tun.east);
    \draw[arr] (tun.north) -- (serv.south);
  \end{tikzpicture}
  \caption{SLO-Tuner treats the server as a black box and tunes concurrency, batch size, and speculative width from tail latency and goodput.}
  \label{fig:arch}
\end{figure}
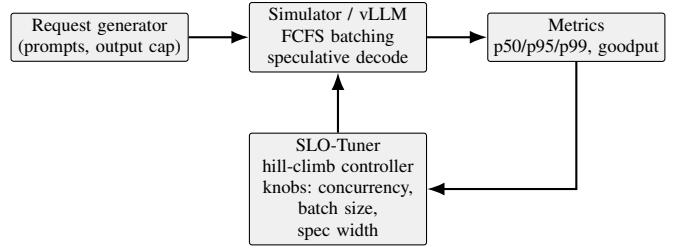

We will step through the overview, controller algorithm, implementation details, simulator, integration, and penalties in turn. \textbf{Overview.} Figure~\ref{fig:arch} shows the architecture. A request generator issues prompts with a maximum output length. Requests are served either by (i) a discrete-event simulator, or (ii) a live vLLM server. A metrics module computes latency percentiles (p50/p95/p99) and goodput from per-request traces. The tuner executes a hill-climb loop: run a short measurement segment, score the configuration, evaluate nearby configurations (neighbors), and move only if the score improves.

\noindent\textbf{Design goals:} (1) \emph{Black-box}: use only standard APIs (e.g., OpenAI-compatible endpoints) and public flags, enabling portability across serving stacks. (2) \emph{SLO-first}: prioritize meeting the p99 SLO over maximizing raw throughput. (3) \emph{Budget-aware}: finish within a bounded number of short segments suitable for real deployments.

Unlike prior tuning work that targets average throughput or requires internal instrumentation, SLO-Tuner optimizes \emph{SLO-satisfying} throughput from black-box measurements and treats speculative decoding as a first-class runtime knob. Conceptually, SLO-Tuner is ``try a few nearby knob settings, keep what improves SLO-satisfying throughput, and back off when p99 worsens.''

For our \textbf{controller algorithm:} We tune a knob vector $K=\{\text{conc},\text{max\_seqs},\text{spec\_tokens},\text{spec\_on/off}\}$, where \texttt{conc} is client concurrency, \texttt{max\_seqs} is \texttt{max\_num\_seqs}, and \texttt{spec\_tokens}/\texttt{spec\_on/off} control speculative decoding. The controller is defined over a small set of \emph{logical knobs} that capture the operator-relevant trade-offs: queueing pressure (concurrency), batch formation (max sequences), and speculation aggressiveness. The simulator exposes speculation in a factorized form---draft width $W$ and verifier cadence $k$---so we can explicitly model acceptance/verification dynamics and stress-test how speculation inflates tail latency~\cite{leviathan2023speculative}. Real serving stacks do not always expose the same factors. For example, vLLM provides public flags such as \texttt{--speculative-model} and \texttt{--num-speculative-tokens}, but it does not expose an independent knob that corresponds exactly to the simulator's verifier cadence $k$~\cite{vllm_docs_spec_decode}.

SLO-Tuner handles this mismatch \emph{by design}: the hill-climb logic is unchanged, and only a thin \emph{adapter} maps the logical knob vector to whatever a given stack exposes. In our live vLLM experiments, we therefore tune a restricted but deployable subset (concurrency, \texttt{max\_num\_seqs}, and speculative token width), without relying on privileged instrumentation or internal scheduler hooks. The controller starts from a baseline configuration and repeatedly evaluates the current setting and a small neighborhood of nearby knob settings. For each candidate $K$, we measure end-to-end latency percentiles (p50/p95/p99) and compute goodput under the target SLO.

 For our implementation, each evaluation uses a short warmup followed by a fixed measurement window; failed/invalid requests (e.g., timeouts or connection errors) are excluded from percentile computation.

For a segment of duration $T$ with per-request latencies $\{\ell_i\}_{i=1}^N$, we define
\begingroup\small
\begin{equation}
  \text{goodput}(K)
  = \frac{1}{T}\sum_{i=1}^N \mathbf{1}\{\ell_i \le \text{SLO}\},
  \label{eq:goodput}
\end{equation}
\endgroup
and write $p_{99}(K)$ for the empirical 99th percentile of $\{\ell_i\}$.

Because some knob increases (e.g., more concurrency or larger batches) also increase resource pressure, we include a simple hardware-intensity proxy:
\begingroup\small
\begin{equation}
\begin{aligned}
  \text{hw\_cost}(K)
  &= w_\text{conc}\,\text{concurrency}
   + w_\text{max}\,\text{max\_num\_seqs} \\
  &\quad + w_\text{spec}\,\text{num\_spec\_tokens},
\end{aligned}
\label{eq:hwcost}
\end{equation}
\endgroup
with $w_\text{conc}{=}w_\text{max}{=}0.01$ and $w_\text{spec}{=}0.02$ in our vLLM runs. These weights are chosen so that p99 SLO violations dominate the score relative to small utilization gains, and we hold them fixed across all experiments to keep comparisons consistent.

The vLLM controller maximizes the score
\begingroup\small
\begin{equation}
\begin{aligned}
  S(K)
  &= \text{goodput}(K)
   - \lambda \max\bigl(0,\ p_{99}(K) - \text{SLO}\bigr) \\
  &\quad - \text{hw\_cost}(K),
\end{aligned}
\label{eq:score}
\end{equation}
\endgroup
where $\lambda$ is the SLO penalty weight (5.0). We set $\lambda$ so that even modest p99 violations outweigh comparable throughput gains, and we keep $\lambda$ fixed across runs. This directly discourages configurations that violate the SLO (bad for users) even if they increase raw utilization.

Neighbor generation is deterministic with fixed step sizes derived from knob ranges: concurrency $\pm2$ within [2,16]; \texttt{max\_num\_seqs} $\pm3$ within [4,16]; speculative width $\pm4$ within [0,16]; plus a speculation on/off toggle. All candidates are clamped to bounds. Each step evaluates the current point and all neighbors; the controller moves to the best neighbor if the score improves by at least $\delta{=}0.02$ or if the current p99 violates the SLO and any neighbor improves the score. Runs stop after a fixed number of iterations (8 in the vLLM setup) without random restarts.

\begingroup\tiny
\begin{algorithm}[!t]
  \caption{SLO-Tuner control loop}
  \label{alg:slo_tuner_loop}
  \begin{algorithmic}
    \REQUIRE Step budget $T$, SLO target $\text{SLO}$, threshold $\delta$
    \STATE Initialize $K \gets K_0$ (concurrency 8, \texttt{max\_num\_seqs} 8, width 8, spec on)
    \STATE $K_{\text{best}} \gets K$, $S_{\text{best}} \gets -\infty$
    \FOR{$t = 1$ to $T$}
      \STATE Run one warmup+measurement segment at $K$
      \STATE Compute $p_{50}(K)$, $p_{95}(K)$, $p_{99}(K)$, $\text{goodput}(K)$ via~\eqref{eq:goodput}
      \STATE Compute $S(K)$ via~\eqref{eq:score}
      \IF{$S(K) > S_{\text{best}}$}
        \STATE $S_{\text{best}} \gets S(K)$; $K_{\text{best}} \gets K$
      \ENDIF
      \STATE $\mathcal{N}(K) \gets$ neighbors of $K$ (steps $\pm2$, $\pm3$, $\pm4$, spec on/off)
      \FORALL{$K' \in \mathcal{N}(K)$}
        \STATE Compute $S(K')$
      \ENDFOR
      \STATE $K^\star \gets \arg\max_{K' \in \mathcal{N}(K)} S(K')$
      \IF{$S(K^\star) - S(K) \ge \delta$ \OR ($p_{99}(K) > \text{SLO}$ \AND $S(K^\star) > S(K)$)}
        \STATE $K \gets K^\star$
      \ENDIF
    \ENDFOR
    \STATE Re-run one segment at $K_{\text{best}}$ and report final metrics
  \end{algorithmic}
\end{algorithm}
\endgroup

Noise handling relies on repeatable segment durations, discarding warmup traffic, and per-step evaluation of all neighbors rather than stochastic sampling. We do not average metrics across steps; instead, the controller keeps the best-so-far configuration and re-measures it at the end to produce the summary. The simulator additionally runs until it reaches both a minimum duration and request count to smooth high-variance segments and uses an exponentially weighted moving average (EMA) of p99 to dampen short-term spikes:
$\hat p_{99}^{(t)}=\beta p_{99}^{(t)}+(1-\beta)\hat p_{99}^{(t-1)}$ with $\beta\in(0,1)$.

\textbf{Simulator.}
We implement a discrete-event simulator to explore trends and stress-test the controller cheaply before tuning a live server. Requests arrive under steady (Poisson) or bursty on/off regimes; arrivals join an FCFS queue; when the server is idle it forms a batch of up to $B$ requests (optionally waiting up to \texttt{max\_wait} ms).

Each batch has two phases. \emph{Prefill} time is driven by the longest prompt in the batch (and a calibrated token rate). \emph{Decode} time depends on the number of active sequences and speculative decoding behavior. Speculation is modeled via a draft width $W$ and verifier cadence $k$: higher draft aggressiveness can reduce verifier steps when acceptance is high, but it can also increase variance through extra verification work when acceptance is low, inflating p99.

Timing parameters are loaded from a configuration file calibrated from measurements, with light noise added for variability. As in the live system, we report p50/p95/p99 and goodput via~\eqref{eq:goodput}. The simulator is not intended to predict absolute kernel-level timing; it is used to capture queueing/batching dynamics and to guide search toward SLO-feasible regions.

\textbf{vLLM integration.}
The controller constructs a vLLM command line with the selected knobs, starts the server, probes \texttt{/v1/models}, and runs an asynchronous client against \texttt{/v1/chat/completions}~\cite{vllm_docs_openai_server}. A 10~s warmup precedes the 30~s measurement window; warmup responses are discarded. Client concurrency, server
\texttt{--max-num-seqs}, and \texttt{--num-speculative-tokens}\allowbreak/\texttt{--speculative-model} are set per segment~\cite{vllm_docs_serve,vllm_docs_spec_decode}. The server is started and stopped for every segment, and the implementation interacts only with vLLM's public API.

\textbf{Cost and SLO penalties.}
The SLO penalty weight $\lambda$ and hardware-cost weights (concurrency, \texttt{max\_num\_seqs}, speculative width) trade off latency violations against resource intensity. Both controllers share the goodput-minus-penalty structure, but the simulator adds draft/verifier-specific cost and a sharper violation penalty (multiplying the violation term by $10\lambda$) so that it retreats aggressively when p99 exceeds the target.

\subsection{Results} 
We present results in three layers. First, the simulator maps the goodput--p99 trade-off and shows how batching and speculative settings affect the SLO surface under steady and bursty load. Second, live vLLM tuning demonstrates that the same hill-climb logic improves p99 and goodput in a real serving stack. Third, targeted ablations isolate how concurrency, batch size, and speculative width individually shape p99 under an explicit SLO.

\textbf{Experimental setup:}

\textbf{Models and workload.} We use TinyLlama/TinyLlama-1.1B-Chat-v1.0, a 1.1B-parameter chat model~\cite{tinyllama2024}, served through vLLM~\cite{kwon2023vllm}. The client replays a single prompt (``You are an expert systems engineer\ldots''), caps responses at 64 tokens, and fixes the arrival process to steady load with closed-loop concurrency. This synthetic, single-prompt workload intentionally removes prompt-mix variability so that we can isolate how serving knobs affect p99 and goodput before tackling more complex traces. In the simulator, prompt and output lengths follow log-normal distributions chosen to mimic a ``large prompt, moderate output'' regime.

\textbf{Metrics and objective.} Latency percentiles are computed over each measurement window, using end-to-end client-observed latency for vLLM. For vLLM runs we target a p99 SLO of 1.2~s; in the simulator we use the same 1.2~s p99 SLO to stress-test batching and speculative decoding decisions under controlled workloads. We reuse the score function from the controller design, with hyperparameters matched to the implementation; goodput dominates the score only when the SLO is met.

\textbf{Simulator configuration.} The simulator search space covers draft width $W\in[1,4]$, verifier width $k\in[2,16]$, batch size $B\in[1,32]$, and optional queueing delay $\texttt{max\_wait}\in[0,50]$~ms. Segments run until at least 1{,}500 completions and 90~s of wall time under a 1.2~s p99 SLO. We bound $W$ to small values because wider drafts add verifier and compute cost with limited upside for a 1.1B model. Prefill/decode token rates and noise parameters follow the configured timing block and drive the discrete-event simulator. Tuning experiments on the simulator use the same hill-climb logic as vLLM but can afford more steps because segments are inexpensive.

\textbf{vLLM configuration.} We use the controller described in Section~2 (same knob bounds, neighbor steps, and scoring). Each control iteration runs 10~s of warmup (discarded) and 30~s of measurement, and the controller runs for 8 iterations to cap tuning time.

\textbf{Hardware and software.} vLLM experiments use a single GPU node requested through SLURM with one NVIDIA L40S GPU (48~GB memory)~\cite{nvidiaL40S_brief}, 8 CPU cores, and 48~GB RAM. The client and server processes are co-located on the node; vLLM is restarted for every segment to ensure knob changes take effect cleanly. We restart vLLM per segment because these flags are not safely hot-swappable in our setup; eliminating restarts is future work. We use vLLM 0.6.1 with FP16 weights and tensor parallelism 1, running on a recent CUDA and driver stack.

\textbf{Simulator experiments.}
We first use the discrete-event simulator to map out the configuration space and stress-test the controller in a controlled environment before touching a real server. All simulator runs use the same timing and workload parameters from \texttt{config/defaults.yaml}, a 1.2~s p99 SLO, and steady or bursty arrival regimes as configured.

\begin{figure}[t]
  \centering
  \subfloat[Baselines (steady).]{\includegraphics[width=0.48\columnwidth]{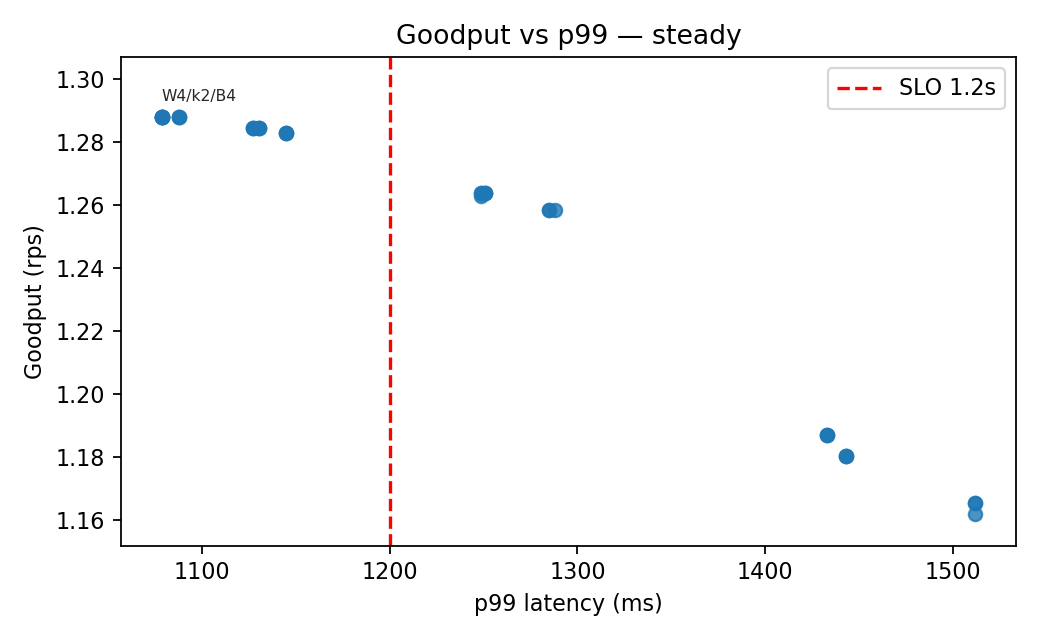}\label{fig:sim-pareto-baselines}}
  \hfill
  \subfloat[Baselines (bursty).]{\includegraphics[width=0.48\columnwidth]{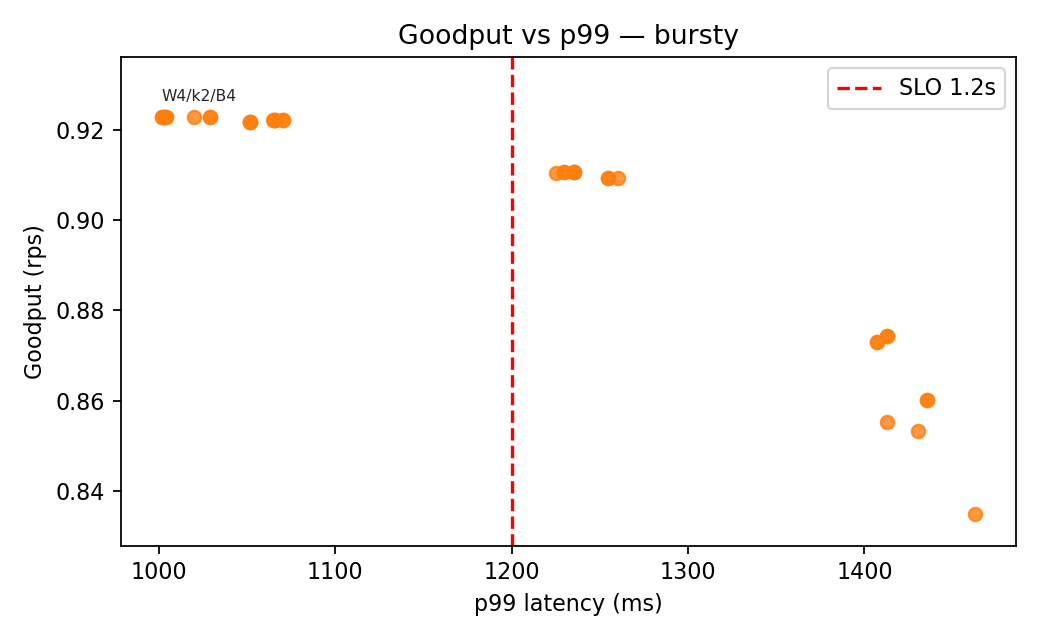}\label{fig:sim-pareto-baselines-bursty}}

  \subfloat[Stress tests (steady).]{\includegraphics[width=0.48\columnwidth]{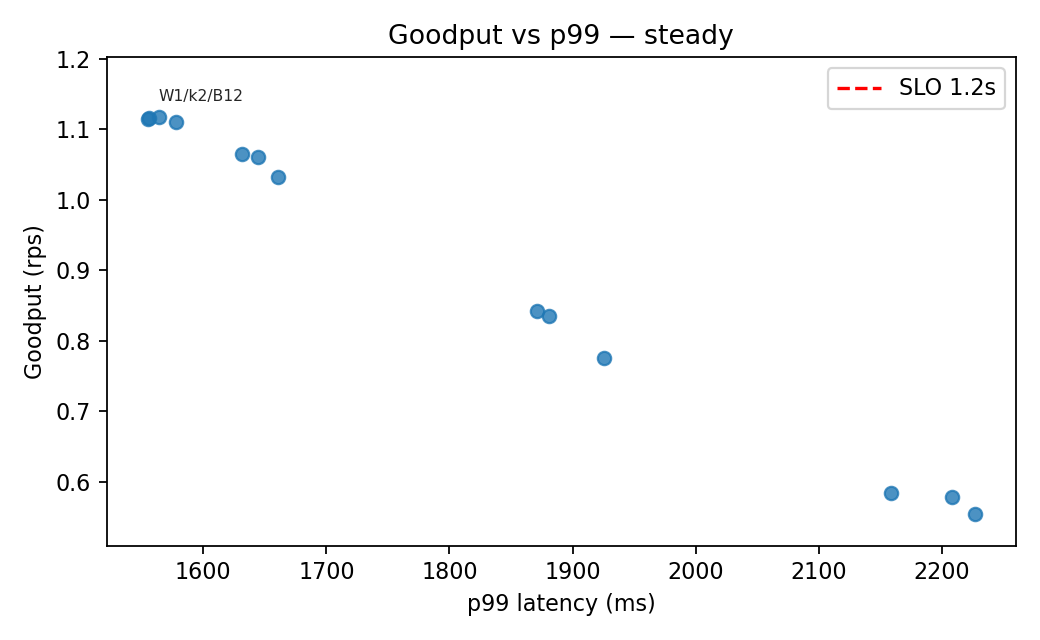}\label{fig:sim-pareto-stress}}
  \hfill
  \subfloat[Stress tests (bursty).]{\includegraphics[width=0.48\columnwidth]{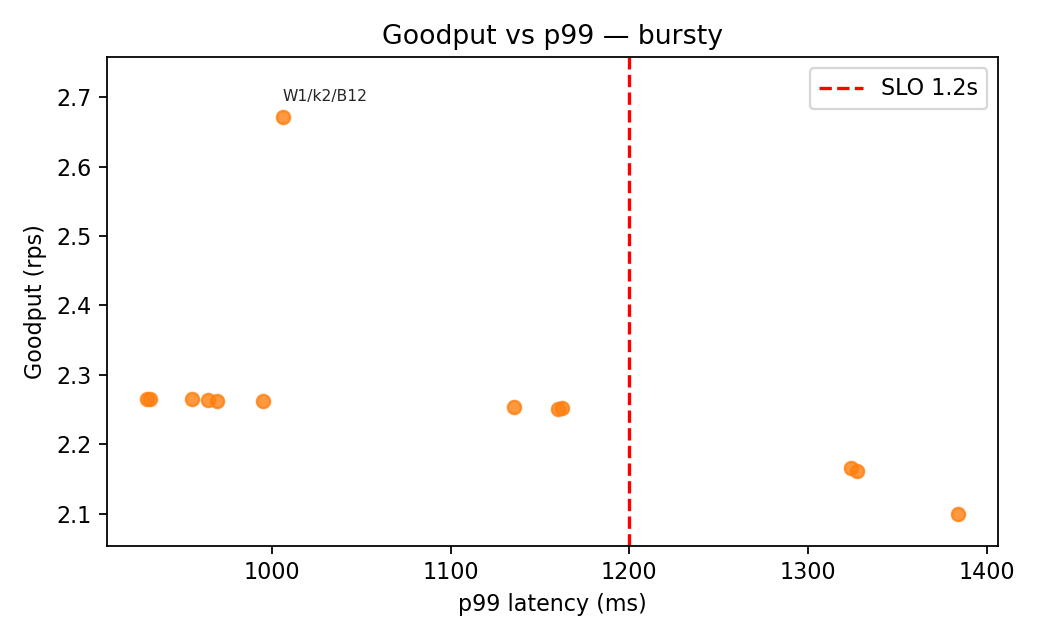}\label{fig:sim-pareto-stress-bursty}}
  \caption{Simulator goodput--p99 tradeoffs (steady vs.\ bursty).}
\end{figure}

\textbf{Baselines.} A baseline grid sweeps $W\in\{1,2,4\}$, $k\in\{2,4,8,12\}$, and $B\in\{4,8,16\}$ under both steady and bursty arrivals. The resulting Pareto front in the steady regime (Fig.~\ref{fig:sim-pareto-baselines}) shows that batching is essential: $B{=}1$ drives p99 to 4.98~s, whereas $B{\geq}2$ can bring p99 near the 1.2~s boundary (and sometimes slightly above it) depending on $W$ and $k$, with $\approx$1.29~rps goodput at near-boundary points. This motivates SLO-aware tuning to push into the SLO-feasible region ($p99 \le 1.2$~s) rather than relying on a single fixed batch setting. However, for small $W$ and conservative $k$, moderate batches achieve high goodput while keeping p99 below 1.2~s; very large batches eventually hurt tails via waiting time. In the bursty regime (Fig.~\ref{fig:sim-pareto-baselines-bursty}), the same pattern holds qualitatively, but points move closer to the SLO boundary as bursts increase queueing, and safe regions shrink to small $W$ and moderate $B$; aggressive $k$ or extreme $B$ values fall far from the frontier in both regimes.

\begin{figure}[t]
  \centering
  \subfloat[Tuning under steady arrivals.]{\includegraphics[width=0.48\columnwidth]{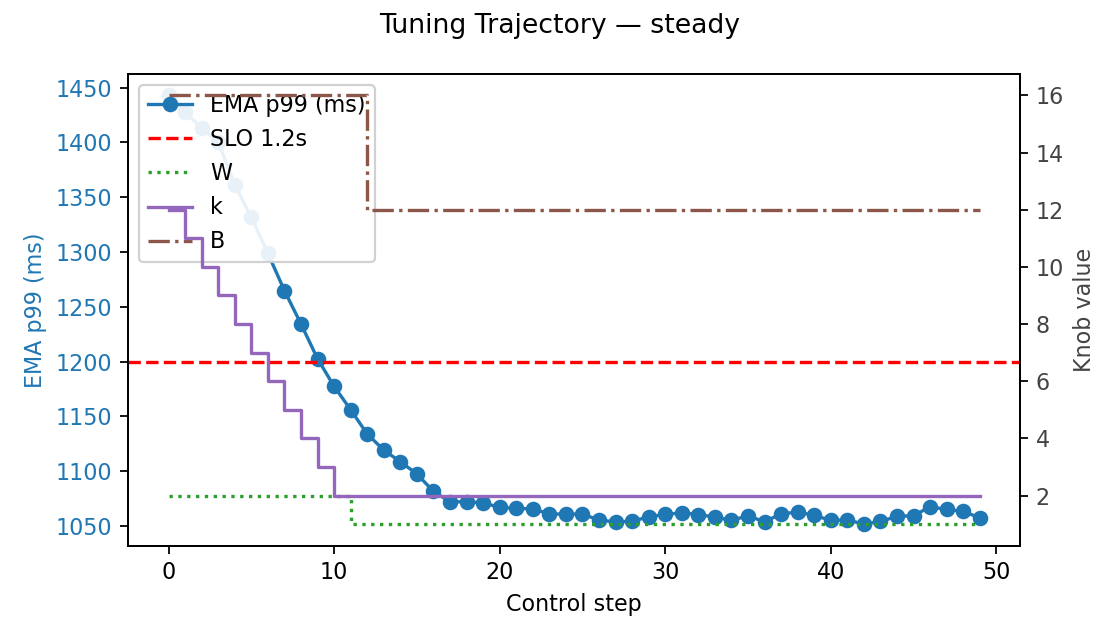}\label{fig:sim-track}}
  \hfill
  \subfloat[Tuning under bursty arrivals.]{\includegraphics[width=0.48\columnwidth]{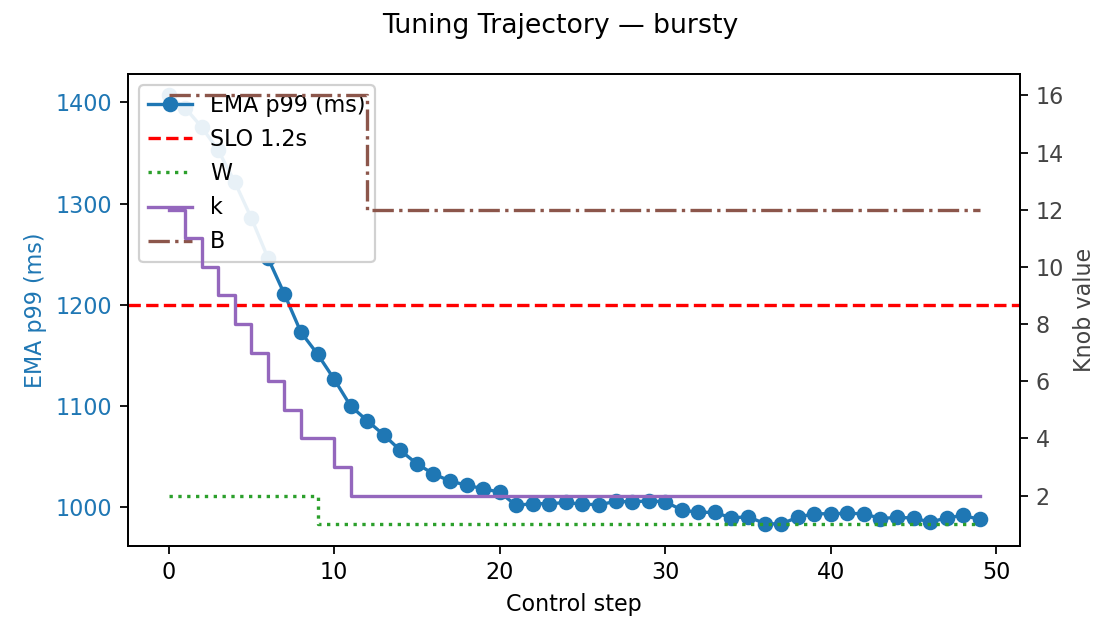}\label{fig:sim-track-bursty}}
  \caption{Simulator hill-climb trajectories.}
\end{figure}

\textbf{Tuning trajectories.} The hill-climb controller runs for six steps per regime, starting from $W{=}2$, $k{=}12$, $B{=}16$ and evaluating one-hop neighbors in $W$, $k$, $B$, and optional queuing delay. In both steady and bursty regimes, the controller quickly moves toward smaller $k$ and slightly reduced $B$, converging to $W{=}1$, $k{=}2$, $B{=}12$ while keeping EMA p99 near the SLO. Figure~\ref{fig:sim-track} shows the steady-regime trajectory: EMA p99 drops sharply over the first few steps as $k$ is reduced, then stabilizes while $B$ is trimmed from 16 to 12, with the controller avoiding oscillation by rolling back when consecutive moves violate the SLO. The bursty trajectory (Fig.~\ref{fig:sim-track-bursty}) is noisier but follows the same qualitative path: the controller first escapes high-$k$ configurations that waste verifier effort and then settles on a small-$W$, moderate-$B$ operating point that keeps p99 under the 1.2~s target despite bursts.

\textbf{Stress tests.} We stress the simulator with longer prompts (mean lengths scaled by $1.5\times$) and a more bursty on/off arrival process while comparing fixed baselines to the tuned profile. In these regimes, aggressive fixed settings that look attractive under steady load---large $B$, high $k$, or wider drafts---often push p99 close to the SLO boundary or beyond it while delivering only marginal goodput gains. The tuned profile, by contrast, typically backs off on speculative width and avoids extreme batches, keeping p99 close to the 1.2~s target while sustaining goodput comparable to the best fixed configurations (Fig.~\ref{fig:sim-pareto-stress} for steady, Fig.~\ref{fig:sim-pareto-stress-bursty} for bursty). This reinforces the need for SLO-aware tuning rather than one-off manual choices tuned to a single benign workload.

\begin{figure}[t]
  \centering
  \subfloat[Ablations (steady load).]{\includegraphics[width=0.48\columnwidth]{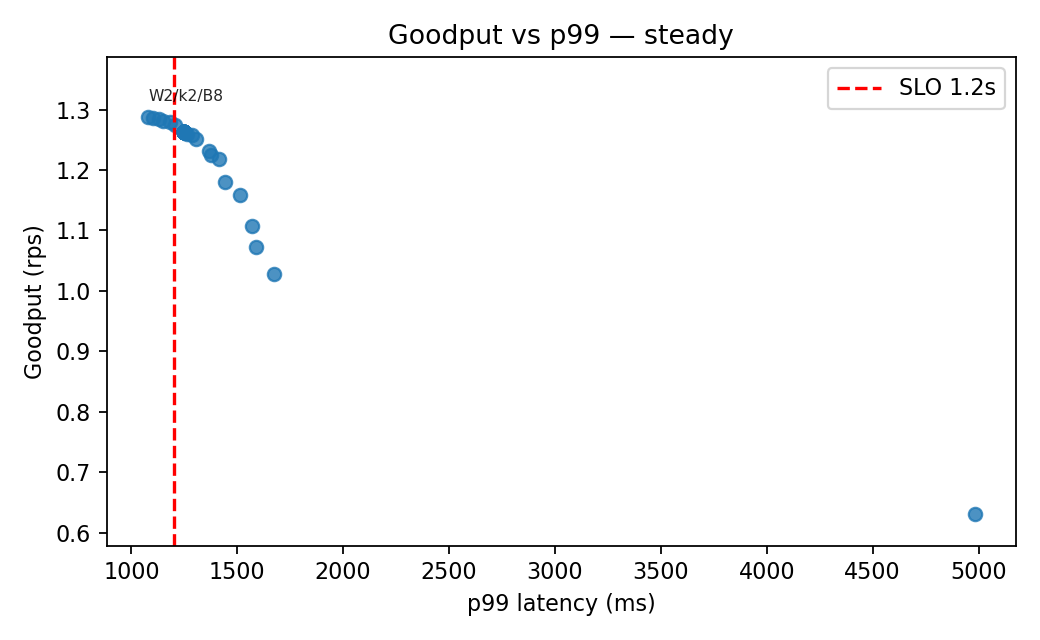}\label{fig:sim-pareto-ablation}}
  \hfill
  \subfloat[Ablations (bursty load).]{\includegraphics[width=0.48\columnwidth]{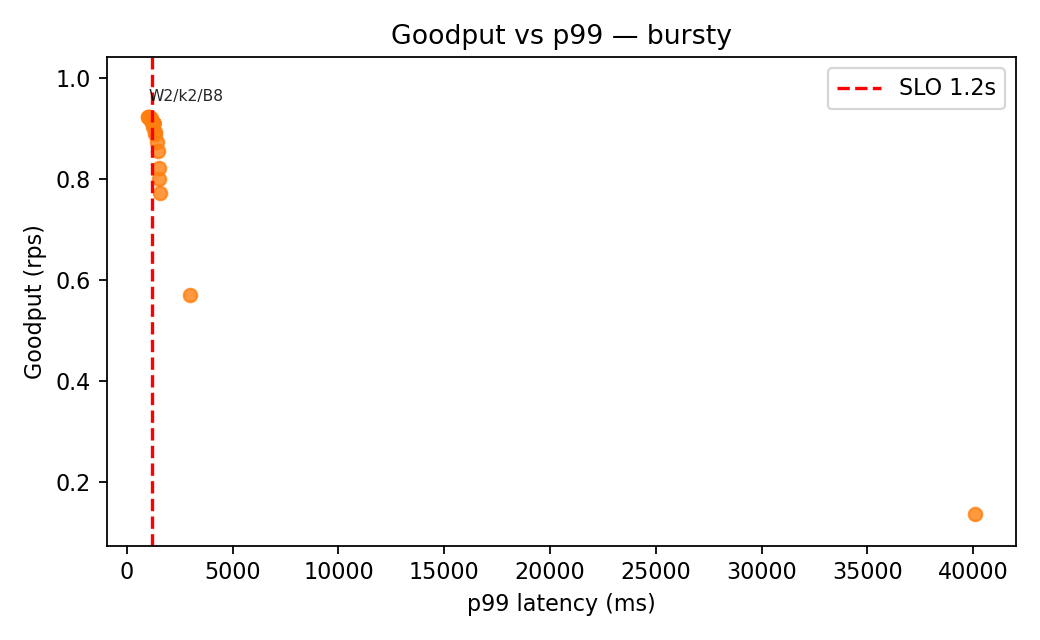}\label{fig:sim-pareto-ablation-bursty}}
  \caption{Simulator ablations over speculative width, verifier cadence, batch size.}
\end{figure}

\textbf{Ablations.} Finally, simulator ablations sweep one knob at a time around a mid-point configuration: fixing $W{=}2$ and $B{=}8$ while varying $k$, fixing $k{=}8$ and $B{=}8$ while varying $W$, and fixing $W{=}2$, $k{=}8$ while sweeping $B$ from 1 to 32. The resulting Pareto plots (Fig.~\ref{fig:sim-pareto-ablation} for steady, Fig.~\ref{fig:sim-pareto-ablation-bursty} for bursty) isolate each trade-off: increasing $k$ steadily raises p99 with little goodput benefit once very sparse verification is reached; widening drafts beyond $W{=}2$ offers negligible goodput gains but can inflate p99 under bursty load; and very small batches underutilize the server while very large batches prolong queueing and hurt tails. Together with the baselines and tuning runs, these ablations support a simple rule of thumb: small speculative width and moderate batches are safest for meeting a tail-latency SLO across both steady and bursty workloads.

We now discuss vLLM live tuning and ablations, followed by simulator vs vLLM alignment followed by our portability check.

\begin{figure}[t]
  \centering
  \subfloat[Online tuning trajectory.]{\includegraphics[width=0.48\columnwidth]{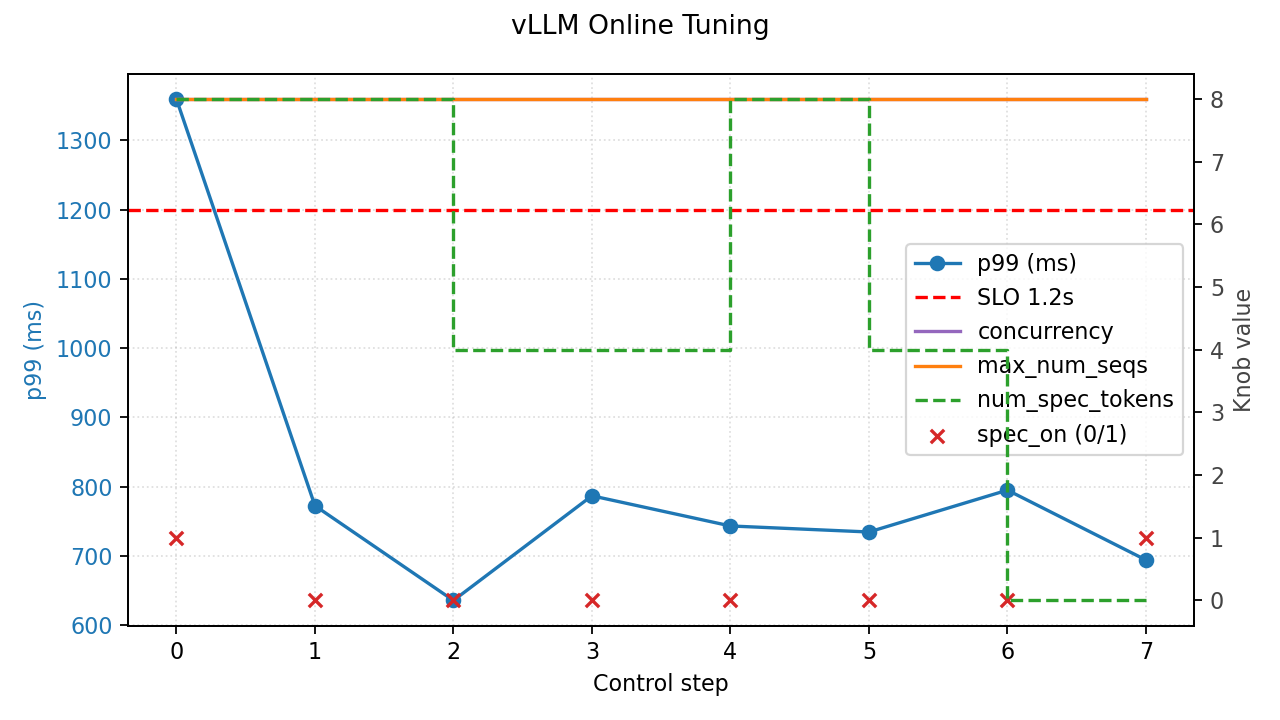}\label{fig:vllm-tune}}
  \hfill
  \subfloat[Concurrency ablation.]{\includegraphics[width=0.48\columnwidth]{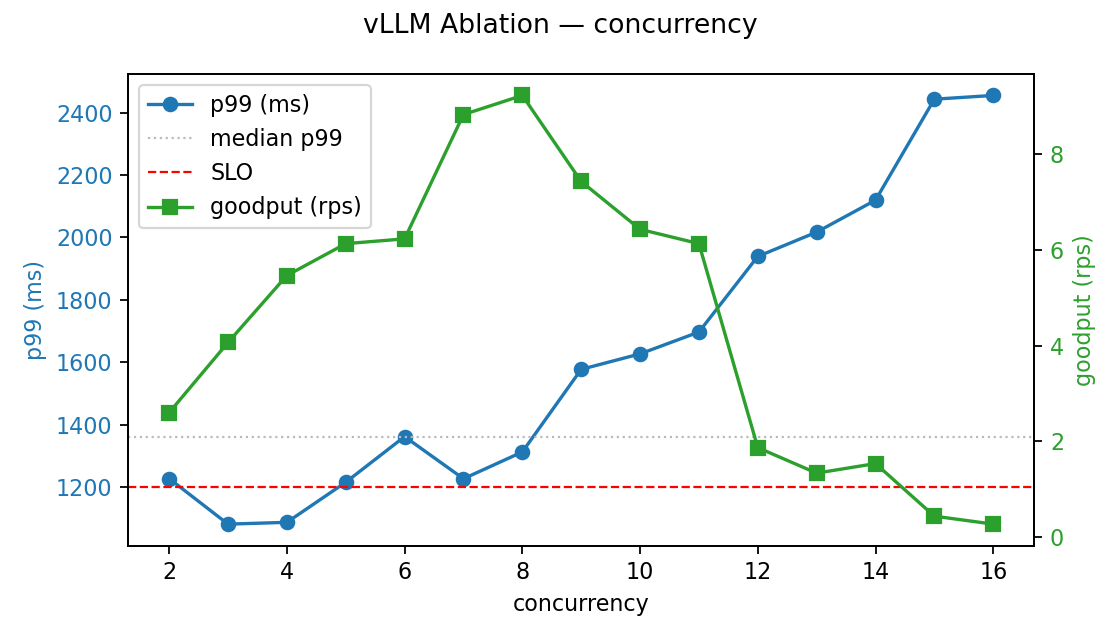}\label{fig:vllm-abl-conc}}

  \subfloat[Batch-size ablation.]{\includegraphics[width=0.48\columnwidth]{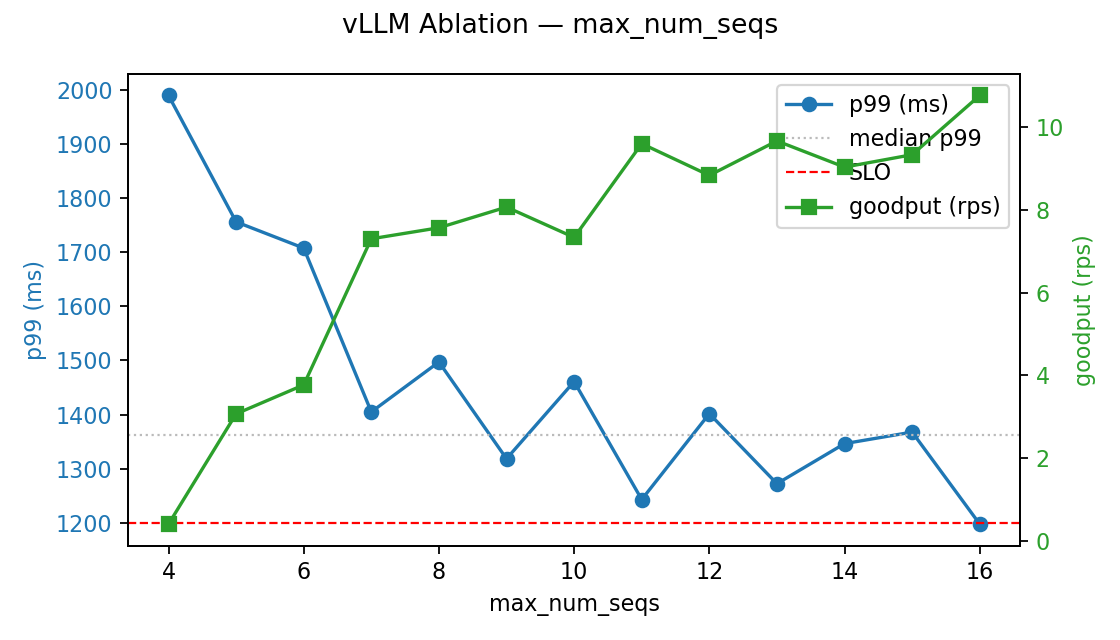}\label{fig:vllm-abl-maxseq}}
  \hfill
  \subfloat[Speculative-width ablation.]{\includegraphics[width=0.48\columnwidth]{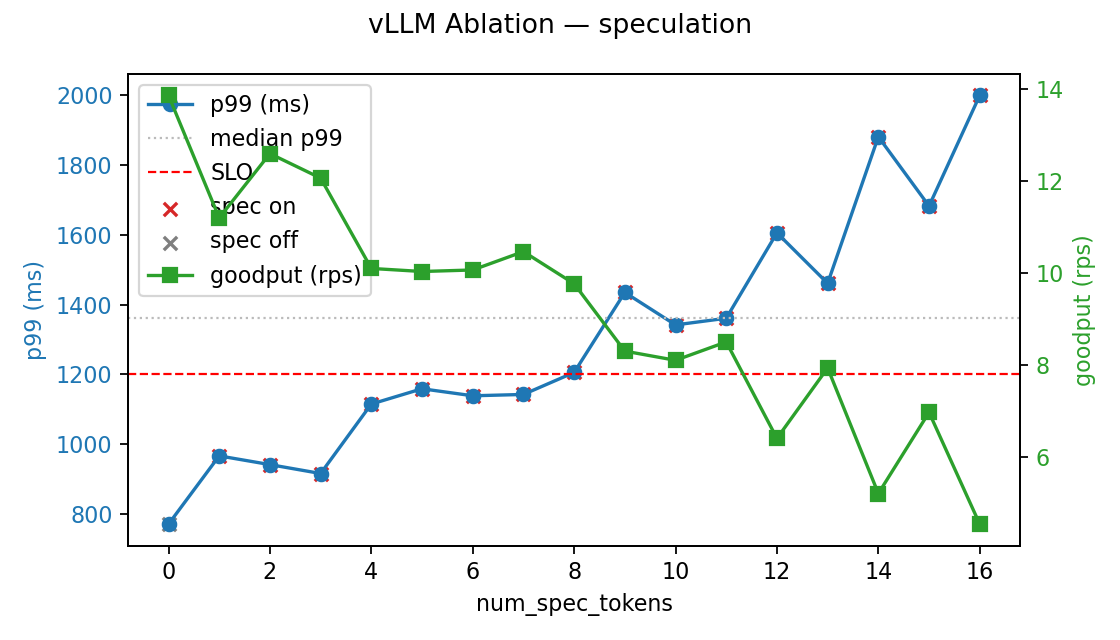}\label{fig:vllm-abl-spec}}
  \caption{vLLM experiments: SLO-Tuner trajectory and ablations over concurrency, batch size, and speculative width.}
\end{figure}

\textbf{vLLM live tuning and ablations.}
On vLLM (TinyLlama), the controller starts from the default-like baseline (concurrency 8, \texttt{max\_num\_seqs} 8, speculative width 8, speculation on), which yields p99 1.36~s and 8.13~rps goodput. The first few hill-climb steps explore toggling speculation and shrinking the draft width while keeping concurrency and batch fixed. Reducing speculative width produces a large jump in goodput and a drop in p99; intermediate widths deliver mixed results, with some segments showing modest improvements and others revealing noise in the SLO surface. In vLLM, setting speculative width to 0 effectively disables speculation (even if the feature flag remains enabled), and the best point (width 0, same concurrency/batch) reaches p99 $\approx$0.7~s and $\approx$15~rps goodput under the 1.2~s SLO. Figure~\ref{fig:vllm-tune} shows the trajectory and knob traces; concurrency and batch remain fixed while speculation moves, confirming that a simple coordinate-wise search over speculative width is enough to escape the default configuration.

Tuning cost is bounded: 8 iterations $\times$ (current + up to 7 neighbors) yields roughly 64 measured segments, each 40~s including warmup, plus a final re-evaluation of the best point. The full run fits in roughly 40--45 minutes of single-GPU time on an L40S, dominated by starting vLLM for each segment. This budget is small enough to run during deployment or configuration changes, and far lower than a full grid search over the same knob ranges.

Concurrency shows a clear elbow: increasing from 2 to 8 threads lifts goodput from 2.6~rps to 9.2~rps while p99 moves from 1.23~s to 1.31~s, but beyond 10 threads p99 exceeds 1.6~s and goodput collapses to 0.27~rps at 16 (Fig.~\ref{fig:vllm-abl-conc}). Raw throughput continues to rise modestly with concurrency, but most requests violate the 1.2~s SLO and thus contribute zero goodput, illustrating why optimizing throughput alone is misleading in an SLO-constrained service.

Batch size (\texttt{max\_num\_seqs}) benefits from larger queues up to a point: raising from 4 to 16 boosts goodput from 0.40~rps to 10.77~rps and lowers p99 from 2.00~s to 1.20~s; the queueing knee appears around 11--13 sequences, where p99 crosses the 1.2~s boundary and goodput begins to plateau or degrade (Fig.~\ref{fig:vllm-abl-maxseq}). For operators, this suggests that moderate batches are sufficient: pushing batch size to the hardware limit delivers only marginal goodput gains while risking higher tail latency if workloads drift.

Speculative width is monotone harmful for this SLO: speculation off (width 0) achieves 13.87~rps at p99 0.77~s, width 8 drops to 9.77~rps at p99 1.21~s, and width 16 falls to 4.53~rps with p99 2.00~s (Fig.~\ref{fig:vllm-abl-spec}). Disabling or shrinking drafts is the only way to respect the 1.2~s p99 target on TinyLlama. This underscores that speculative width should be treated as a tunable knob whose optimal setting depends on the SLO and workload, not as an always-beneficial optimization.

\textbf{Simulator vs.\ vLLM alignment.}
Both simulator and vLLM show the same ordering on speculative width: small widths (or off) keep p99 low and goodput high, while wider drafts raise p99 without commensurate goodput gains. For comparable widths, simulated p99 is 1.25--1.36~s versus 0.77--1.21~s on vLLM; the magnitudes differ ,but goodput and p99 trend in the same direction as width grows. The simulator caps width at 4, so it underestimates the sharp degradation seen at 16 speculative tokens in vLLM; aligning extreme-width behavior is future work.

\begin{table}[t]
  \centering
  \renewcommand{\arraystretch}{0.9}
  \caption{Headline vLLM results (TinyLlama on vLLM, 1.2~s p99 SLO).}
  \label{tab:headline}
  \begin{tabular}{@{}lcc@{}}
    \toprule
    Knob setting & p99 (s) & Goodput (rps) \\
    \midrule
    Baseline: spec width 8, spec on & $\approx$1.36 & $\approx$8.1 \\
    Tuned: spec width 0, spec on & $\approx$0.70 & $\approx$15.0 \\
    \bottomrule
  \end{tabular}
\end{table}

\begin{figure}[t]
  \centering
  \includegraphics[width=0.32\columnwidth]{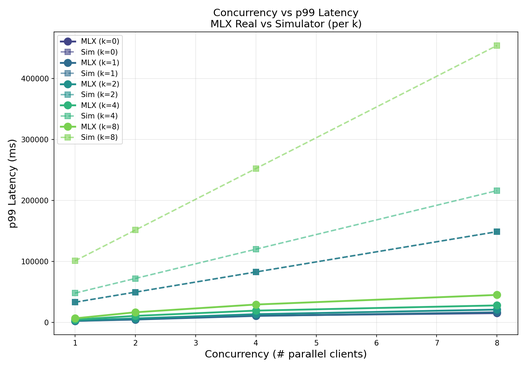}\hfill
  \includegraphics[width=0.32\columnwidth]{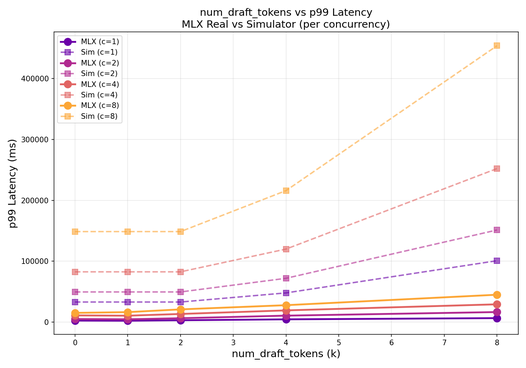}\hfill
  \includegraphics[width=0.32\columnwidth]{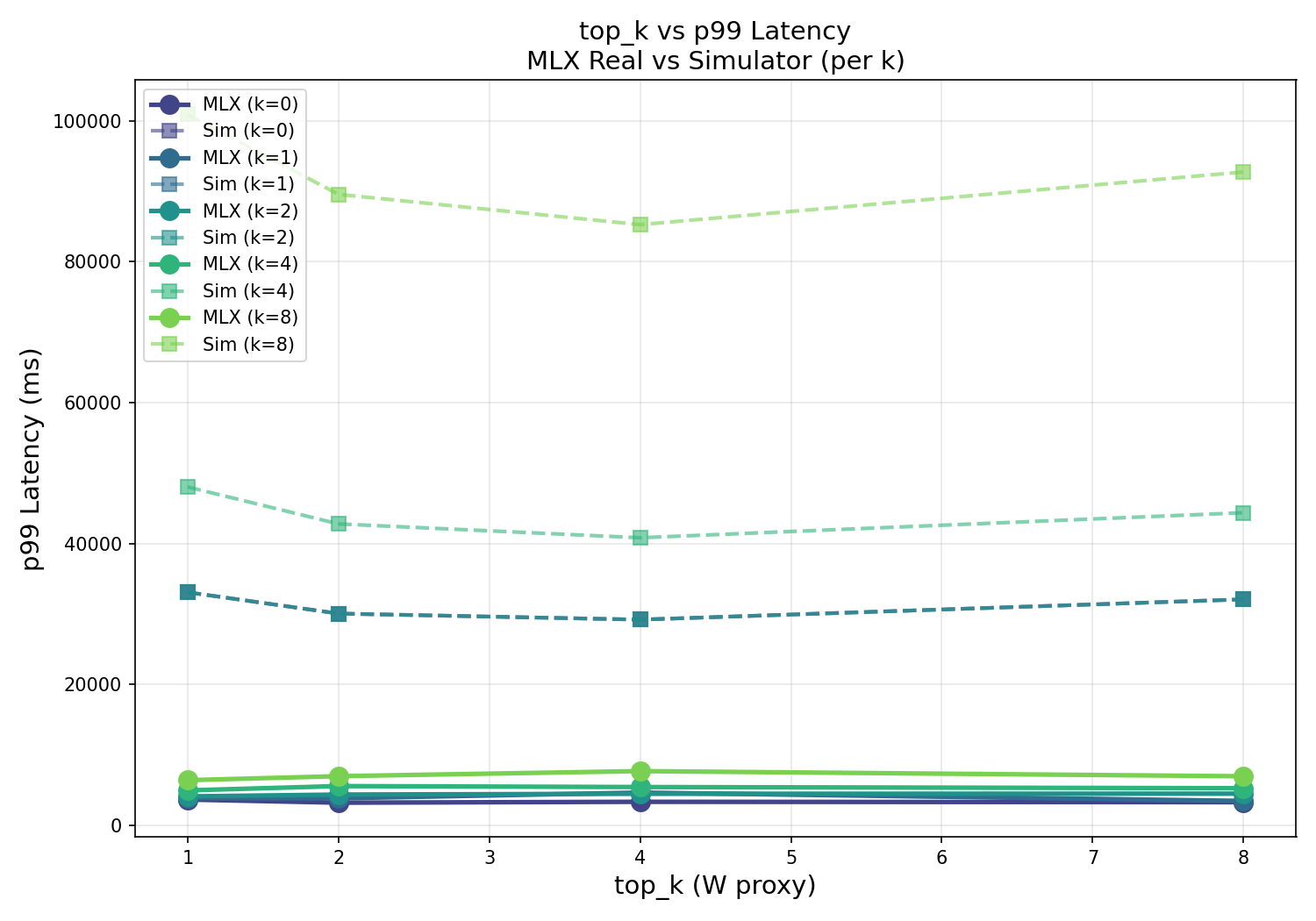}
  \caption{MLX vs.\ simulator: concurrency, lookahead $k$, and top-$k$ width. Dashed=sim, solid=MLX.}
  \label{fig:mlx_val}
\end{figure}

\textbf{Portability check: MLX on Apple Silicon.}
To sanity-check portability beyond NVIDIA GPUs, we ran a small companion evaluation on Apple Silicon using MLX~\cite{hannun2023mlx} with a Qwen-4B target and a Qwen-0.6B draft model. We sweep three operator-facing knobs—client concurrency, speculative lookahead ($k$), and draft width (proxied by top-$k$)—and compare p99 trends to the simulator (Fig.~\ref{fig:mlx_val}). While the simulator shows a clear calibration offset in \emph{absolute} latency, it matches the \emph{directional} response to knob changes: increasing concurrency degrades p99, and more aggressive speculation can inflate tail latency. This is exactly the role we need from the simulator in our workflow: it provides cheap stress-testing and trend-correct guidance toward SLO-feasible regions, while the online controller remains necessary to adapt to hardware-specific saturation points that a static timing model cannot predict.

\section{Discussion and Limitations} \label{sec:limitations}

\noindent\textbf{Discussion.}
Speculative width is not ``the wider the better'': on TinyLlama, wide drafts inflate p99 and reduce goodput, so the controller often shrinks or disables speculation despite its throughput appeal. This complements prior work that focuses on throughput and cost, and highlights that speculative decoding should be tuned jointly with batch size and concurrency when tight SLOs are present. For practitioners, the key lesson is that enabling speculative decoding with a large width can silently violate SLOs even when average latency and throughput look attractive.

The black-box approach worked end-to-end on vLLM without code changes, suggesting it can retarget other stacks (e.g., TGI, Triton, MLX)~\cite{hf2023tgi,nvidia2022triton,hannun2023mlx} with minimal integration effort. Because SLO-Tuner interacts only through standard APIs and command-line flags, it can be deployed alongside existing monitoring and autoscaling infrastructure rather than replacing them. In a production deployment, SLO-Tuner could be run during off-peak hours or on canary replicas, then roll out the best configuration to the rest of the fleet.

\noindent\textbf{Limitations and threats to validity.}
Our work is preliminary. We evaluate 3 relatively small models (TinyLlama-1.1B, Qwen-0.6B, Qwen-4B) on a single GPU node, using a synthetic single-prompt workload and steady closed-loop arrivals. Real deployments exhibit mixed prompt and output lengths, heterogeneous request types, and time-varying arrival patterns (including diurnal cycles and burstiness). While our simulator approximates some variability, it does not model multi-tenant interference, multi-GPU scheduling effects, or cluster-level admission control, all of which can materially affect tail latency.

Our search method is first-order hill climbing (8 fixed steps) rather than bandits or Bayesian optimization. This provides predictable tuning overhead but can get trapped in local optima if the performance surface is non-convex. We also use a fixed scoring function; other SLOs, hardware, or workloads may require re-tuning these hyperparameters. More adaptive scoring and alternative search strategies remain important directions.

Experimentally, we report a single vLLM tuning trajectory rather than a distribution over many runs with different seeds or starting points, limiting our ability to characterize variance and robustness. We also do not include random search or other gradient-free optimizers under the same segment budget; such comparisons would strengthen the evaluation but require additional GPU time.

Finally, SLO-Tuner may underperform on non-stationary or multi-tenant workloads where the SLO surface shifts faster than the controller can react, or where upstream mechanisms (e.g., admission control, rate limiting, caching) interact with the tuned knobs. Strict operational budgets could also make repeated server restarts untenable, motivating designs that update knobs online without restarts.

\noindent\textbf{Conclusion.}
SLO-Tuner is a practical, black-box controller for p99-aware LLM serving with speculative decoding. It couples a discrete-event simulator with a live vLLM integration to tune concurrency, batching, and speculative width through public APIs. On TinyLlama, it halves p99 latency (1.36~s $\rightarrow$ 0.7~s) while nearly doubling goodput (8~rps $\rightarrow$ 15~rps) under a 1.2~s SLO. Looking forward, tightening simulator-to-system calibration, extending to larger models and more realistic workloads, handling multi-tenant settings, and validating across additional serving stacks and SLO targets are key next steps.

\section{Responsible AI, Trust, and Sustainability} \label{sec:responsible}
Responsible AI has looked at various considerations for judging an LLM to be acceptable to organizations which decide to employ them. Standard considerations include transparency, accuracy, fairness and bias~\cite{benk2025bridging,cohen2019trusted}, advocated in order to promote trustworthiness. In addition to novel technical solutions to advance improved human--AI relationships~\cite{cheng2021sociallyresponsible}, the use of more effective communication methods has also been promoted. Some very interesting work has been proposing the introduction of factsheets in order to work towards certifying that a particular AI system is effective for its users, without unexpected negative outcomes~\cite{arnold2019factsheets}. Consideration of the system performance of LLMs has not traditionally been a part of factsheet proposals but we suggest here that it should be. Our work has explored ways in which the systems running LLMs can be improved. One question which arises is whether there are any significant concerns, if these performances do not improve. If the factsheets and insights into whether an AI system can be considered responsible fail to pass judgment on the system performance, what particular negatives may emerge which would then begin to impact some of the standard concerns for Responsible AI?

One central issue is whether the AI systems that we construct can be trusted. LLMs in particular can be prone to hallucinations or the data used to train them may have hidden biases that compromise possible employment towards decision making that affects the lives of individuals~\cite{weidinger2022taxonomy}. As system performance degrades, we observe that this may open the door to organizations adjusting their parameters in order for the systems to complete their tasks and yield their results. And yet in so doing the datasets being used may be reduced, and unless that step is taken carefully, additional problems with bias and fairness may emerge. Another option may be to set aside the need for transparency and explanation, since pushing the system to produce results more quickly may come at the expense of this~\cite{kambhampati2020challenges}. We feel that these concerns are very important, aligned with views presented in position papers~\cite{varshney2015datascience} that clarify the importance of robust system design when systems dictate crucial outcomes. Another important consideration which arises as part of Responsible AI is sustainability and we notice that this has been integrated into a recent significant effort on Responsible AI~\cite{montreal2018declaration}. Additional focus on this concern with the increased use of our AI systems in organizations today is also advisable. Efforts such as ours which seek to enable LLMs to run more efficiently will help towards that cause.

The general value of research into sustainable AI has also recently been underscored in venues such as COMPASS with valued work that also promotes inclusivity~\cite{atinafu2025pixlift}. One outcome of the work in our paper can be to elicit new attention for this element to also somehow be integrated into Responsible AI factsheets. In the end, systems performance matters, first to continue to promote beneficial adoption of LLMs within organizations and secondly to mitigate additional harms when shortcuts to improve performance may arise. We note that others have already considered trustworthiness at all stages of the life cycle of AI systems~\cite{toreini2020trust} and thus system performance issues being considered is very relevant. 
The work presented in this paper serves to bolster the argument that making system performance of LLMs matter more to users is not only important but also feasible.

In our work, practitioners are able to see how system throughput is performing with the current implementation, based on choice of LLM and settings of parameters; Table 5 shows one example. Some  current work on sustainability ~\cite{atinafu2025pixlift} also provides users with quantitative values for degradation of outcomes based on percentage of disadvantaged population. Promoting continued insights in these areas of concern to those adopting LLMs today is what is what we would like to see, in the future.

\section*{AI-Generated Content Acknowledgement}

Generative AI tools were used solely for proofreading and minor grammatical edits in parts of the manuscript, including some table text. The research content and findings are entirely the authors’ work.

\bibliographystyle{IEEEtran}
\bibliography{references}

@inproceedings{kwon2023vllm,
  author    = {Kwon, Woosuk and others},
  title     = {Efficient Memory Management for Large Language Model Serving with {PagedAttention}},
  booktitle = {SOSP},
  year      = {2023}
}

@inproceedings{leviathan2023speculative,
  author    = {Leviathan, Yaniv and others},
  title     = {Fast Inference from Transformers via Speculative Decoding},
  booktitle = {ICML},
  year      = {2023}
}

@inproceedings{chen2019ottertune,
  author    = {Chen, Sheng and others},
  title     = {OtterTune: Automatic Database Management System Tuning},
  booktitle = {SIGMOD},
  year      = {2019}
}

@article{dean2013tail,
  author  = {Dean, Jeffrey and others},
  title   = {The Tail at Scale},
  journal = {Communications of the ACM},
  volume  = {56},
  number  = {2},
  pages   = {74--80},
  year    = {2013}
}

@inproceedings{delimitrou2014quasar,
  author    = {Delimitrou, Christina and others},
  title     = {Quasar: Resource-Efficient and QoS-Aware Cluster Management},
  booktitle = {ASPLOS},
  year      = {2014}
}

@inproceedings{alipourfard2017cherrypick,
  author    = {Alipourfard, Omid and others},
  title     = {CherryPick: Adaptively Unearthing the Best Cloud Configurations for Big Data Analytics},
  booktitle = {NSDI},
  year      = {2017}
}

@techreport{nvidia2022triton,
  author      = {{NVIDIA}},
  title       = {Triton Inference Server},
  institution = {NVIDIA},
  year        = {2022},
  note        = {Software}
}

@techreport{hf2023tgi,
  author      = {{Hugging Face}},
  title       = {Text Generation Inference},
  institution = {Hugging Face},
  year        = {2023},
  note        = {Software}
}

@inproceedings{cai2024medusa,
  author    = {Cai, Tianle and others},
  title     = {Medusa: Simple {LLM} Inference Acceleration Framework with Multiple Decoding Heads},
  booktitle = {ICML},
  year      = {2024}
}

@inproceedings{ankner2024hydra,
  author    = {Ankner, Zachary and others},
  title     = {Hydra: Sequentially-Dependent Draft Heads for Medusa Decoding},
  booktitle = {COLM},
  year      = {2024}
}

@inproceedings{gan2021sinan,
  author    = {Zhang, Yanqi and others},
  title     = {Sinan: {ML}-Based and QoS-Aware Resource Management for Cloud Microservices},
  booktitle = {ASPLOS},
  year      = {2021}
}

@software{hannun2023mlx,
  author = {Hannun, Awni and others},
  title  = {{MLX}: Efficient and Flexible Machine Learning on Apple Silicon},
  year   = {2023},
  note   = {Software}
}

@misc{montreal2018declaration,
  author = {{Universit{\'e} de Montr{\'e}al}},
  title  = {The Montr{\'e}al Declaration for a Responsible Development of Artificial Intelligence},
  year   = {2018},
  note   = {Website. Accessed 2025-12-14}
}

@article{arnold2019factsheets,
  author  = {Arnold, Matthew and others},
  title   = {FactSheets: Increasing Trust in {AI} Services through Supplier's Declarations of Conformity},
  journal = {IBM Journal of Research and Development},
  volume  = {63},
  number  = {4/5},
  pages   = {6:1--6:13},
  year    = {2019},
  doi     = {10.1147/JRD.2019.2942288}
}

@article{cheng2021sociallyresponsible,
  author  = {Cheng, Lu and others},
  title   = {Socially Responsible {AI} Algorithms: Issues, Purposes, and Challenges},
  journal = {J. Artif. Intell. Res.},
  year    = {2021}
}

@inproceedings{weidinger2022taxonomy,
  author    = {Weidinger, Laura and others},
  title     = {Taxonomy of Risks Posed by Language Models},
  booktitle = {FAccT},
  year      = {2022}
}

@inproceedings{benk2025bridging,
  author    = {Benk, Michaela and others},
  title     = {Bridging the Knowledge Gap: Understanding User Expectations for Trustworthy {LLM} Standards},
  booktitle = {AAAI},
  year      = {2025}
}

@misc{varshney2015datascience,
  author       = {Varshney, Kush R. and others},
  title        = {Data Science of the People, for the People, by the People: A Viewpoint on an Emerging Dichotomy},
  howpublished = {D4GX},
  year         = {2015}
}

@inproceedings{cohen2019trusted,
  author    = {Cohen, Robin and others},
  title     = {Trusted {AI} and the Contribution of Trust Modeling in Multiagent Systems},
  booktitle = {AAMAS (Blue Sky)},
  year      = {2019}
}

@inproceedings{toreini2020trust,
  author    = {Toreini, Ehsan and others},
  title     = {The Relationship Between Trust in {AI} and Trustworthy Machine Learning Technologies},
  booktitle = {FAccT},
  year      = {2020}
}

@inproceedings{atinafu2025pixlift,
  author    = {Atinafu, Yonas and others},
  title     = {PixLift: Accelerating Web Browsing via {AI} Upscaling},
  booktitle = {COMPASS},
  year      = {2025}
}

@article{kambhampati2020challenges,
  author  = {Kambhampati, Subbarao and others},
  title   = {Challenges of Human-Aware {AI} Systems: {AAAI} Presidential Address},
  journal = {AI Magazine},
  volume  = {41},
  number  = {3},
  pages   = {3--17},
  year    = {2020}
}

@misc{googleSRE_slos,
  author = {{Google}},
  title  = {The Art of {SLO}s},
  year   = {2025},
  note   = {Google SRE Workbook. Accessed 2025-12-18}
}

@misc{vllm_docs_openai_server,
  author = {{vLLM Project}},
  title  = {OpenAI-Compatible Server (vLLM Documentation)},
  year   = {2025},
  note   = {Docs. Accessed 2025-12-18}
}

@misc{vllm_docs_spec_decode,
  author = {{vLLM Project}},
  title  = {Speculative Decoding (vLLM Documentation)},
  year   = {2025},
  note   = {Docs. Accessed 2025-12-18}
}

@misc{vllm_docs_serve,
  author = {{vLLM Project}},
  title  = {Serving (vLLM Documentation)},
  year   = {2025},
  note   = {Docs. Accessed 2025-12-18}
}

@techreport{nvidiaL40S_brief,
  author      = {{PNY Technologies}},
  title       = {NVIDIA {L40S} Product Brief},
  institution = {PNY Technologies},
  year        = {2023},
  note        = {Product brief. Accessed 2025-12-18}
}

@article{tinyllama2024,
  author  = {Zhang, Peiyuan and others},
  title   = {TinyLlama: An Open-Source Small Language Model},
  journal = {arXiv},
  volume  = {abs/2401.02385},
  year    = {2024}
}

@inproceedings{cheng2025scoot,
  title     = {{SCOOT}: {SLO}-Oriented Performance Tuning for {LLM} Inference Engines},
  author    = {Cheng, Ke and others},
  booktitle = {Proceedings of the ACM Web Conference 2025 (WWW '25)},
  year      = {2025},
  pages     = {829--839},
  publisher = {ACM},
  doi       = {10.1145/3696410.3714930}
}

\end{document}